# UID2021: An Underwater Image Dataset for Evaluation of No-reference Quality Assessment Metrics

GUOJIA HOU*, YUXUAN LI, and HUAN YANG,

College of Computer Science & Technology, Qingdao University, China

KUNQIAN LI,

College of Engineering, Ocean University of China, China

ZHENKUAN PAN,

College of Computer Science & Technology, Qingdao University, China

Achieving subjective and objective quality assessment of underwater images is of high significance in underwater visual perception and image/video processing. However, the development of underwater image quality assessment (UIQA) is limited for the lack of comprehensive human subjective user study with publicly available dataset and reliable objective UIQA metric. To address this issue, we establish a large-scale underwater image dataset, dubbed UID2021, for evaluating no-reference UIQA metrics. The constructed dataset contains 60 multiply degraded underwater images collected from various sources, covering six common underwater scenes (i.e. bluish scene, bluish-green scene, greenish scene, hazy scene, low-light scene, and turbid scene), and their corresponding 900 quality improved versions generated by employing fifteen state-of-the-art underwater image enhancement and restoration algorithms. Mean opinion scores (MOS) for UID2021 are also obtained by using the pair comparison sorting method with 52 observers. Both in-air NR-IQA and underwater-specific algorithms are tested on our constructed dataset to fairly compare the performance and analyze their strengths and weaknesses. Our proposed UID2021 dataset enables ones to evaluate NR UIQA algorithms comprehensively and paves the way for further research on UIQA. Our UID2021 will be a free download and utilized for research purposes at: https://github.com/Hou-Guojia/UID2021.

## 1 INTRODUCTION

The quality of underwater images or videos plays a critical role in marine scientific research and ocean engineering, such as sea life monitoring, object detection and tracking. However, underwater captured images often suffer from low contrast, blurring, distorted color, noise, and other degradations due to the absorption and scattering effects when light travels under the water. To address these problems, many methods [1-3] have been developed for underwater image enhancement and restoration (UIER). In these methods, subjective observation is always used to evaluate their performance. Although subjective evaluation can give a reliable result, it is expensive, laborious, and non-automatic. Therefore, it is important to achieve an effective objective metric for image quality assessment (IQA). Generally, existing objective IQA metrics can be classified into three categories: full-reference (FR), reduced-reference (RR), and no-reference (NR) algorithms, depending on the availability of a reference image. For underwater scenarios where a reference image is usually not available, the no-reference IQA metrics are highly desired. But unfortunately, not all existing natural image quality metrics can be effectively applied to evaluate the underwater image because they fail to correlate well with subjective perception. So far, the efficient NR metrics specifically designed for underwater image quality assessment (UIQA) are UCIQE [4], UIQM [5], CCF [6] and FDUM [7], which constrains the development of underwater image/video enhancement and restoration. In order to figure out the performance of the state-of-the-art (SOTA) NR IQA metrics for underwater image evaluation, it is necessary to conduct a comprehensive evaluation of the modern NR metrics on the available dataset. However, all the available natural images datasets for IQA research cover a diverse set of in-air images, they are not applicable to judge and optimize UIER algorithms.

* The research work is partially supported by the National Natural Science Foundation of China (No. 61901240, 61906177), the Natural Science Foundation of Shandong Province, China (No. ZR2019BF042, ZR2019BF034), China Scholarship Council, China (No. 201908370002), and the China Postdoctoral Science Foundation (No. 2017M612204).
Authors' addresses: G. Hou (corresponding author), Y. Li, H. Yang and Z. Pan, College of Computer Science & Technology, Qingdao University, No. 308 Ningxia Road, Qingdao, China, 266071; emails: hgj2015@qdu.du.cn, lyx776239423@gmail.com, cathy_huanyang@hotmail.com, zkpan@126.com; K. Li, College of Engineering, Ocean University of China, No. 238 Songling Road, 266100, Qingdao, China; email: likunqian@ouc.edu.cn.

Recently, some researchers developed several underwater benchmark datasets for evaluating underwater image enhancement and restoration methods. In [8-10], three real-world underwater image datasets namely UIEB, RUIE, and SQUID are respectively constructed for studying the SOTA UIER algorithms qualitatively and quantitatively. Besides, several synthetic underwater image datasets are accordingly developed, such as 3D TURBID dataset [11], UID-LEIA dataset [12], UIDD dataset [13], and SUID dataset [14]. Since these proposed synthetic datasets contain reference images, which creates a possibility for FR evaluation of existing UIER technologies. More recently, several real-world underwater image datasets for evaluating NR UIQA algorithms are successively proposed, such as UWIQA [7], UOQ [15], and UEIQA [16]. However, these datasets contain a very limited number of images and cover fewer challenging scenes, which limits their application.

In this paper, we aim to construct a new underwater image dataset namely UID2021, and further provide a comprehensive evaluation of no-reference quality assessment metrics for underwater images. The main contributions of this work include:

(i) A large-scale underwater image enhancement dataset is established containing 60 multiply degraded underwater images and 900 corresponding quality improved versions. In our UID2021, the selected 60 raw underwater images cover six common underwater scenes with different types of degradation. Moreover, 900 images are produced by employing fifteen SOTA algorithms, well designed for underwater image enhancement and restoration.

(ii) The proposed dataset also contains the corresponding subjective mean opinion scores (MOS) from 52 observers. With the constructed UID2021, we further conduct a comprehensive study of nine NR quality assessment metrics.

(iii) The evaluation results can provide a better understanding of the strengths and limitations of current NR IQA metrics and point to the top-performing NR metrics for UIQA. Also, it can suggest a new research direction, so as to facilitate researchers to select appropriate NR metrics to evaluate their newly developed underwater image enhancement and restoration algorithms.

(iv) Our constructed UID2021 can also be regarded as a benchmark to guide the optimization process of developing new NR IQA algorithms for specific underwater images.

The rest of the paper is organized as follows. The existing IQA databases including the natural in-air image dataset and underwater image dataset are reviewed in Section 2. Section 3 presents a review of the popular NR IQA metrics. Section 4 describes the generation of the proposed UID2021 dataset in detail. In Section 5, we present the evaluation results, as well as the associated discussions. Finally, Section 6 concludes the paper.

Table 1. Comparison of existing eleven in-air IQA datasets.

| Database | IVC | LIVE | MICT | TID2008 | CSIQ | MDIQ | TID2013 | CID2013 | MDID | MDIVL | WED |
|---|---|---|---|---|---|---|---|---|---|---|---|
| Ref. | [26] | [17] | [18] | [19] | [27] | [23] | [24] | [20] | [21] | [25] | [22] |
| Database Type | STD | STD | STD | STD | STD | MTD | MTD | MTD | MTD | MTD | MTD |
| No. of Source Images | 10 | 29 | 14 | 25 | 30 | 15 | 25 | 8 | 20 | 10 | 4744 |
| No. of Distorted Images | 185 | 779 | 196 | 1700 | 866 | 405 | 3000 | 480 | 1600 | 750 | 94880 |
| Distortion Type | 4 | 5 | 2 | 17 | 6 | 3 | 24 | 12-14 | 5 | 2 | 4 |
| Distortion Level | 5 | 5 | 6 | 4 | 4-5 | 4 | 5 | N/A | 4 | 12-14 | 5 |
| Number of Subjects | 15 | 20-29 | 16 | 838 | 25 | 37 | 971 | 188 | 192 | 12 | N/A |
| Subjective Test Method | DSIS | SS-HR | SS | PC | N/A | SS-HR | PC | ACR-DR | PCR | SS | N/A |
| Subjective Data Type | DMOS | DMOS | RAW | MOS | DMOS | DMOS | MOS | RAW | MOS | MOS | N/A |
| Score Range | 1-5 | 0-100 | 1-5 | 0-9 | 0-1 | 0-100 | 0-9 | 0-100 | 0-8 | 1-100 | N/A |

## 2 REVIEW OF EXISTING IQA DATASETS

### 2.1 In-air Image Dataset

Over the last couple of decades, many publicly available image databases have been reported to evaluate the FR and NR image quality assessment algorithms, such as Laboratory Image and Video Engineering (LIVE) [17], Multimedia Information and Communication Technology (MICT) [18], Tampere Image Database 2008 (TID 2008) [19], Camera Image Database (CID2013) [20], Multiply Distorted Image Database (MDID) [21], Waterloo Exploration Database (WED) [22] and so on. We pick out some detailed information from eleven publically available in-air IQA databases, as presented in Table 1. As shown in Table 1, IQA databases can be categorized into two types: single type of distortion (STD) and multiple types of distortions (MTD). Different from STD, MTD contains multiple types of distortions of each distorted image. Among them, the STD image databases have originally attracted the greatest attention. Recently, several multiply distorted image databases have been developed such as Multiply Distorted Image Quality (MDIQ) [23], Tampere Image Database 2013 (TID2013) [24], CID2013, MDID, and Multiple Distorted IVL (MDIVL) database [25]. For collecting the subjective data, Single Stimulus (SS) and Pair Comparison (PC) are two of the most widely used schemes. Their benchmark evaluation ratings have different score ranges. In the subjective evaluation, Image and Video Communication (IVC) [26], LIVE, Categorical Subjective Image Quality (CSIQ) [27], and MDIQ databases provide subjective data in form of difference mean opinion score (DMOS). Differently, in TID 2008, TID 2013, MDID, and MDIVL databases, the mean opinion score (MOS) was computed for subjective testing. It is necessary to note that WED is the current largest one for in-air image quality assessment, which contains 4744 clear reference images and 94880 distorted images. Because of the large-scale of WED, it is extremely difficult to collect the MOSs via subjective testing. Instead, they utilized three criteria to perform a systematic evaluation of 20 well-known IQA algorithms. Although these existing datasets cover a wide variety of in-air images, they contain few underwater images, which limits the development of underwater enhancement and restoration algorithms. To fit this gap, some researchers transfer their attention to constructing underwater image datasets.

Table 2. Comparison of different underwater image datasets.

| Dataset | No. of Images | Resolution | Image Generation | Usage | No. of Subjects | Subjective Test Method | Subjective Data Type |
|---|---|---|---|---|---|---|---|
| TURBID [28] | 570 | 4000×3000 | Adding mike to simulate different level of turbidity. | UIER | N/A | N/A | N/A |
| UIEB [8] | 950 | 225×225 to 2180×1447 | Real underwater images taken under natural light, artificial light | UIER | N/A | N/A | N/A |
| RUIE [9] | 4230 | 400×300 | Real underwater images captured by a multi-view imaging system underwater seawater. | UIER | N/A | N/A | N/A |
| NWPU [29] | 6240 | 1600×1200 | Using different level of turbidity water, light conditions, view distances. | UIER | N/A | N/A | N/A |
| UOQ [15] | 216 | 512×512 | Using five image enhancement algorithms | UIQA | N/A | SS | MOS |
| UWIQA [7] | 890 | 225×225 to 2180×1447 | Images collected from the UIEB | UIQA | 21 | SS | MOS |
| UEIQA [16] | 240 | 1280×720 | Using five image enhancement algorithms | UIQA | 18 | SS | MOS |
| UID2021 | 960 | 512×384 | Using fifteen underwater image enhancement and restoration algorithms | UIQA | 52 | PCS | MOS |

## 2.2 Underwater Image Dataset

Currently, to the best of our knowledge, there are two kinds of underwater image datasets used to test the performance of underwater image enhancement or restoration algorithms (UIER) and underwater image quality assessment algorithms, respectively. The details of these two kinds of underwater image datasets can be found in Table 2. With regard to the performance validation of UIER algorithms, Felipe et al. [28] proposed an underwater turbidity images dataset, called TURBID. All the images in TURBID are captured in a 1000 liters water tank, and the level of turbidity is controlled by successively adding milk into the water tank. Afterward, Ma et al. [29] used real turbidity lake water to simulate the real underwater situation and proposed an underwater turbidity image dataset, namely NWPU. NWPU contains 6240 underwater images of 40 objects, which are captured under 6 different levels of turbidity, 4 light conditions, and 6 different distances. More recently, Li et al. [8] proposed a large-scale underwater image enhancement benchmark (UIEB), which includes 890 real underwater images taken under natural light, artificial light, and a mixture of natural and artificial lights. In the same year, Liu, et al. [9] constructed a real-world underwater enhancement (RUIE) dataset, with over 4000 underwater images captured by a multi-view imaging system under seawater. RUIE consists of three subsets: underwater image quality set (UIQS), underwater color cast set (UCCS), and underwater higher-level task-driven set (UHTS), which are used to validate the capability of UIER algorithms to improve image visibility, correct color cast, and the effectiveness from the aspect of high-level underwater tasks.

The aforementioned underwater image datasets focus on testifying the performance of UIER algorithms and do not provide the corresponding MOS values, which are not suitable for evaluating UIQA algorithms. Due to the outstanding performance of FR IQA algorithms, some researchers attempted to adopt FR IQA metrics in evaluating the quality of underwater images to remedy the shortage of inconsistent results between subjective evaluation and some NR IQA metrics. Nevertheless, the existing classical FR IQA algorithms are not available on account of lacking reference images in the underwater environment. To overcome this issue, several synthetic underwater image datasets have been developed. In [13], Li et al. proposed an underwater image synthesis algorithm based on the underwater imaging physical model to generate an underwater images degradation dataset (UUID). In the same year, Hou et al. [14] constructed a large-scale synthetic underwater image dataset (SUID) by utilizing hierarchical searching and red channel prior algorithms to acquire the underwater background light and transmission map from the real-world underwater image. The SUID contains 900 degraded images, covering a diverse set of different turbidity types and degradation levels.

However, there still exists a gap between synthetic and real-world underwater images. Additionally, since the reference images are not available in underwater scenes, NR algorithms are still the first choice of objective quality assessment for underwater images. Nevertheless, all the aforementioned underwater image datasets are not developed for evaluating NR algorithms. Until fairly recently, several underwater images datasets specifically designed for evaluating underwater NR algorithms are emerging. In [15], Wu et al. built an underwater optical image quality database, called UOQ. UOQ consists of 36 typical underwater images with the size of $512 \times 512$, and 180 enhanced images, covering 10 categories of degradation types. Additionally, Yang et al. [7] built an underwater image quality assessment (UWIQA) dataset by directly utilizing 890 images from UIEB as the source images and conducting a subjective quality evaluation to obtain MOS, but the MOS values of the UWIQA dataset are concentrated in 10 discrete scales (from 0.1 to 1), which may be not accurate enough. Moreover, Guo et al. [16] constructed a UEIQA dataset, consisting of 40 source images captured by a remotely operated underwater vehicle and corresponding 200 enhanced images by five image enhancement algorithms. However, all the 40 source images in UEIQA were captured on a sea cucumber farm, indicating that it only covers a limited set of underwater scenes.

The current underwater datasets pay more attention to validating the performance of UIER algorithms rather than UIQA algorithms. Even though several datasets for testing UIQA algorithms are emerging, they usually contain monotonous content and limited scenes, few degradation characteristics, and insufficient data, which makes it difficult or even impractical to fairly evaluate underwater NR algorithms, and hinders the development of new NR algorithms as well. To address these limitations, a new real-world underwater image database is highly desired to provide the benchmark for UIQA algorithms.

## 3 REVIEW OF NR IQA METIRCS

In this section, we will present a brief review of some well-known NR IQA algorithms which are usually used for evaluating the performance of underwater image enhancement and restoration methods. NR IQA methods aim to evaluate the quality of a digital image without depending on any reference information, and they are also known as blind IQA (BIQA) methods. In the last decade, natural scene statistics (NSS) based algorithms have been widely used for NR IQA. Based on the NSS, Moorthy and Bovik [30] designed a new two-step framework for no-reference image quality assessment, namely BIQI, which can assess the image quality across no distortion-specific categories. Mittal et al. [31] also proposed the blind/referenceless image spatial quality evaluator (BRISQUE) based on the spatial domain NSS because of its higher computational efficiency. In BRISQUE, they extracted the natural scene statistics by using local divisive normalization. Afterward, Mittal et al. [32] further developed a natural image quality evaluator (NIQE) without training any human opinion scores. Unlike BRISQUE, NIQE algorithm only considered some image patches with high contrast for fitting NSS features. Not long ago, Saad et al. [33] proposed a novel no-reference image quality assessment method BLIIDNS-II by using an NSS model of discrete cosine transform (DCT) coefficients. BLIIDNS-II utilizes a lower-dimensional feature space and a simpler single-stage framework, which is computationally appealing. With the fast development of deep neural networks (DNN), designing data-driven algorithms to tackle this challenge becomes possible. Ma et al. [34] presented a multi-task end-to-end optimized network (MEON) for NR IQA. MEON is performed by training in two sub-networks: a distortion type identification network and a quality prediction network relying on pre-trained early layers. Motivated by meta-learning, Zhu et al. [35] proposed a new NR IQA metric called MetaIQA by employing bilevel gradient optimization to learn the shard prior knowledge model of various distortions from plenty of NR IQA tasks, and then fine-tune the prior model to obtain the target quality model. In [36], Wu et al. proposed an IQA-oriented CNN method, namely CaHDC, inspired by the hierarchical perception mechanism in the human visual system (HVS). Benefiting from the hierarchical degradation concatenation and the end-to-end optimization, CaHDC can better learn the nature of quality degradation and accurately predict the image quality.

Since haze and blur are two special types of problems of underwater images, some specially designed NR metrics [37-42] are employed to assess the performance of defogging and deblurring algorithms. Among them, Narvekar and Karam [37] first proposed an NR perceptual-based blur metric depending on the notion of just noticeable blur (JNB). JNB can accurately predict the relative amount of blurriness with different content. Afterward, they presented an improved blurriness metric [38] to estimate the probability of detecting blur based on the cumulative probability of blur detection (CPBD) and JNB. Li et al. [39] developed a blind image blur evaluation (BIBLE) metric using discrete orthogonal moments. They divided the gradient image into equal-size blocks and computer their Tchebichef moments to characterize image shape. The proposed BIBLE algorithm not only can generate accurate image blur scores, but also achieves highly consistent with subjective evaluations. In 2015, Choi et al. [40] designed an NR fog density prediction model called fog aware density evaluator (FADE). As applications, FADE can accurately predict the fog density benefiting from its dependence on NSS and fog-aware statistical features. Liu et al. [41] proposed a new NR metric called PSQA, which is based on the analysis of pre-attention and spatial dependency's influence on the perception of distortion. In [42], Yan et al. firstly trained an image distortion classifier by employing the state-of-the-art Inception-ResNet-v2 neural networks. Then, based on the distortion classification, they developed a novel NR metric namely DIQM to characterize the image quality in a targeted manner.

Unlike in-air images, underwater images encounter the effect of serious absorption and scattering. Some IQA metrics for in-air images are not applicable to evaluate underwater image quality. To address this challenge, some authors focus on developing specific non-reference metrics for UIQA. In 2015, Yang and Sowmya [4] designed an underwater color image quality evaluation (UCIQE) algorithm to qualify the non-uniform color cast, blurring, and low-contrast by a linear combination of these three components. The following year, Panetta et al. [5] presented an NR underwater image quality measure (UIQM) inspired by the properties of the HVS. The UIQM algorithm comprises three attribute measures in terms of colorfulness measure, sharpness measure, and contrast measure. Afterward, Wang et al. [6] proposed an imaging-inspired metric for underwater color image quality assessment, called CCF, which is weighted with a linear combination of colorfulness, contrast and fog density. CCF can quantify the color loss, blurring, and foggy, which are caused by absorption, forward scattering and backward scattering, respectively. Likewise, Yang et al. [7] presented a reference-free underwater image quality assessment metric namely FDUM in frequency domain by combining the colorfulness, contrast and sharpness. In addition, some authors perform specific applications such as edge detection [43-44], local feature point matching [45-46], and image segmentation [47], to assess their enhanced and restored results.

Because of the fact that most existing no-reference (NR) metrics are not specially designed for underwater image quality assessment. The above-mentioned in-air IQA metrics are also often employed for UIQA. However, to our best knowledge, there lacks a systematic study of their performance on evaluating the underwater image enhancement and restoration algorithms. Actually, some in-air NR objective metrics show a poor correlation with subjective evaluations, which cannot accurately predict underwater image quality. A detailed description of the nine well-known and widely used NR IQA metrics being evaluated on the proposed database will be given in Section 5.

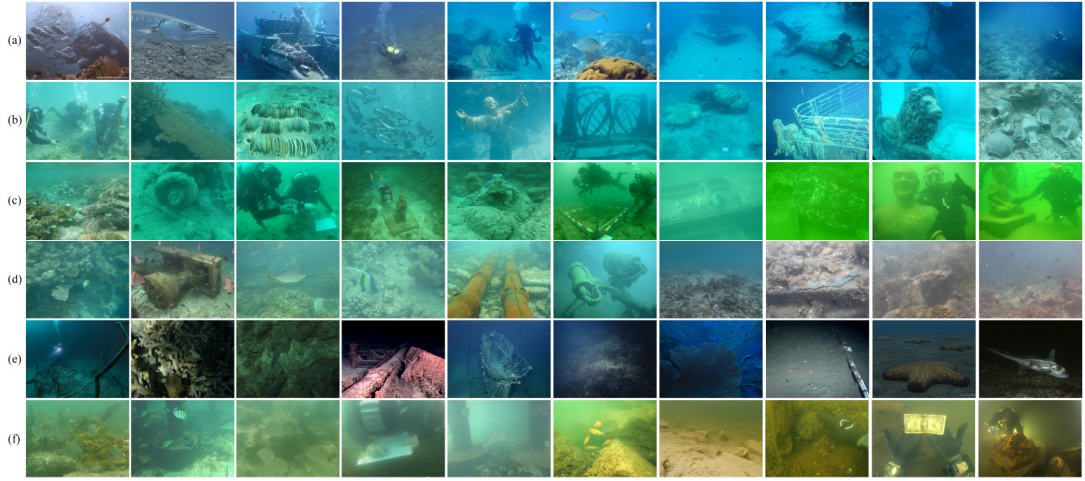

Fig. 1. Source images with different challenging scenes. (a) Blueish scene, (b) blue-green scene, (c) greenish scene, (d) hazy scene, (e) low light scene, and (f) turbid scene.

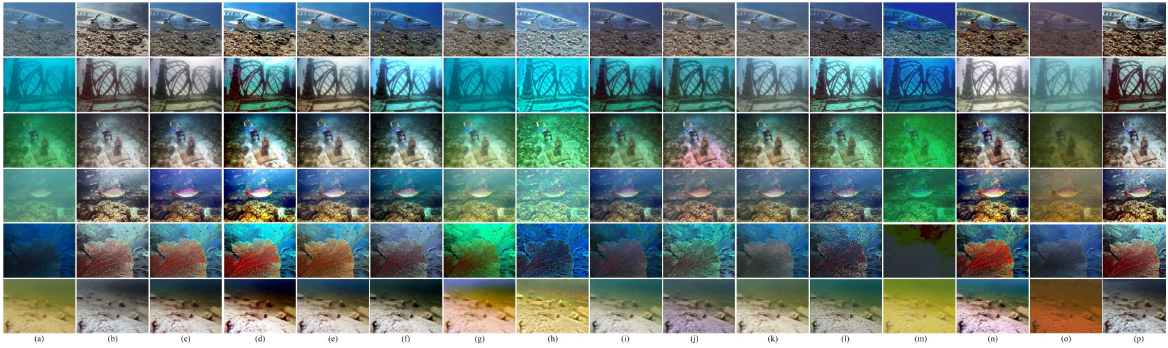

Fig. 2. Source images and their enhanced or restored versions. From left to right are: (a) source images and enhanced or restored results by using (b) Bayesain-retinex, (c) CBF, (d) CHS, (e) GLN-CHE, (f) HP, (g) IBLA, (h) L$^2$UWE, (i) RCP, (j) TS, (k) Ucolor, (l) UNTV, (m) UTV, (n) UWB-VCSE, (o) UWCNN, and (p) VR, respectively.

## 4   UID2021 DATASET

After systematically summarizing previous work, we find that the lack of comprehensive human subjective user study with publicly available datasets and reliable objective UIQA metrics makes it difficult to better understand the true performance of UIER algorithms. Generally, an in-air image database is constructed by dozens of reference images and lots of generated distorted images. However, as for underwater scenes, it is challenging to collect real-world underwater/clear image pairs. Unlike in-air image databases, in our UID2021, the degraded underwater image is regarded as the source image, and the corresponding generated images are produced by applying different enhancement or restoration algorithms. In what follows, we will introduce the constructed database UID2021 in detail, including source image collection and enhanced or restored image generation.

Table 3. Details of the six subsets with different challenge scenes.

| Subset | Scene | Reference images | Total images | Description | Example images |
|---|---|---|---|---|---|
| 1 | Bluish | 10 | 160 | In underwater environment, the decay of light is related to the wavelength of the color. When traveling through water, the red light decays fastest than green and blue wavelengths because of its largest wavelength. Therefore, underwater captured images always appear to have blusih, blue-green or greenish tones. | 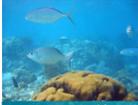 |
| 2 | Blue-Green | 10 | 160 | | 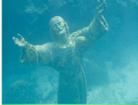 |
| 3 | Greenish | 10 | 160 | | 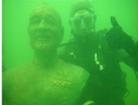 |
| 4 | Hazy | 10 | 160 | The particles suspended underwater lead to images suffer from haze. | 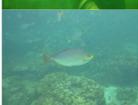 |
| 5 | Low-light | 10 | 160 | In deep water, there's not enough light, and artificial light is needed. | 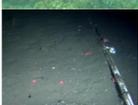 |
| 6 | Turbid | 10 | 160 | Underwater mud makes underwater images turbid. | 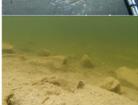 |

## 4.1 Source Images Selection

To our knowledge, the selection of source images plays an important role for UIQA database. Therefore, the selected source images in our UID2021 cover six common types of underwater scenes (i.e., blueish scene, greenish scene, blue-green scene, hazy scene, turbid scene, and low light scene) with different image contents. The details of these six scenes are given in Table 3. The 60 source images are selected from several popular underwater databases [8-10] and some public websites. The selected underwater images are in different size because they are picked from various public real-world image datasets. We crop them into the uniform size of 512×384 to make sure that they can be completely and clearly displayed on the monitor. All the source underwater images are presented in Fig. 1.

## 4.2 Dataset Creation

In our UID2021, based on the 60 source underwater images under different challenging scenes, 900 enhanced or restored images are generated by using fifteen SOTA underwater image enhancement and restoration algorithms (see Table 4 for more details). These existing SOTA methods are specially designed for improving the quality of underwater images from different aspects of contrast enhancement, dehazing, deblurring, color correction, and so on. Due to the limited space, we present one image of each scene accompanied by their fifteen enhanced or restored versions for sample, as shown in Fig. 2. As we can clearly see from Fig. 2, the enhanced or restored versions of the source images cover a broad range of image qualities, assuring that they can be regarded as a benchmark to evaluate the NR algorithms.

Table 4. Details of the fifteen underwater enhancement or restoration methods.

| Category | Method | Public implementation code |
|---|---|---|
| Enhancement | Bayesian-retinex[48] | https://github.com/zhuangpeixian/Bayesian-Retinex-Underwater-Enhancement |
| | CBF[47] | https://github.com/bilityniu/underimage-fusion-enhancement |
| | CHS[49] | Implemented by ourselves |
| | HP[51] | https://github.com/Hou-Guojia/HP |
| | L$^2$UWE[53] | https://github.com/tunai/l2uwe |
| | TS[54] | https://xueyangfu.github.io/paper/2017/ISPACS/code.zip |
| | UWB-VCSE[46] | https://github.com/Hou-Guojia/UWB-VCSE |
| | VR[58] | https://xueyangfu.github.io/projects/icip2014.html |
| Restoration | IBLA[52] | https://github.com/ytpeng-aimlab/Underwater-Image-Restoration-based-on-Image-Blurriness-and-Light-Absorption |
| | RCP[43] | https://github.com/agaldran/UnderWater |
| | UNTV[56] | https://github.com/Hou-Guojia/UNTV |
| | UTV[57] | https://github.com/Hou-Guojia/UTV |
| Deep learning | GLN-CHE[50] | https://xueyangfu.github.io/projects/spic2020.html |
| | Ucolor[55] | https://github.com/Li-Chongyi/Ucolor |
| | UWCNN[13] | https://github.com/saeed-anwar/UWCNN |

Table 5. Five aspects of subjective quality assessment.

| Protocols | Description |
|---|---|
| Color distortion | Underwater image usually suffers from server color distortion. |
| Contrast distortion | The backward scattering effect reduces the contrast of underwater images. |
| Texture distortion | The forward scattering result in blurry underwater images, and the texture information of underwater images are seriously lost. |
| Visibility | Visibility in underwater images is often poor due to reduced contrast, poor lighting conditions, and other problems. |
| Recognizable foreground | Recognizable foreground is important for high-level underwater tasks. |

## 4.3 Subjective Image Quality Assessment

Using reliable subjective evaluation to represent the ground truth is the most significant component for obtaining MOS values for a database. Various subjective methods are employed to provide the reliability of perceptual quality evaluations, such as double stimulus continuous quality scale (DSCQS) scheme, single stimulus (SS) scheme, pair comparison (PC) scheme. DSCQS method is successfully introduced in the IVC database because it is appropriate for evaluating a small number of images. SS method is a widely used scheme, which has been exploited in LIVE, MICT, MDIQ, and MDIVL databases. In this method, the observers are asked to absolutely rate only one image each time. In fact, it is often difficult for observers to assign a score to an image. In many cases, the observers find that they give an improper score when they later evaluate another image, but many subjects are reluctant to change their previous scores. Unlike SS, PC scheme does not require the observers to assign an absolute score of an available image. Instead, they are only asked to choose the better one of two images, which helps observers easefully make a decision. This method is successfully utilized in TID2008 and TID2013 databases. Nevertheless, we find that it is also difficult for observes to compare two quality-indistinctive images. Fortunately, in [21], Sun et al. proposed a novel subjective scheme dubbed pair comparison sorting (PCS). PCS allows observers to make an "equal" decision besides "greater" or "less". Actually, these subjective schemes are not viable when the number of images is tremendous, such as WED database [22] which contains 4744 source images and 94880 distorted images. In [22], Ma et al. presented three criteria, terms as the pristine/distorted image discriminability test, the listwise ranking consistency test, and the pairwise preference consistency test, to perform a subjective evaluation of image quality.

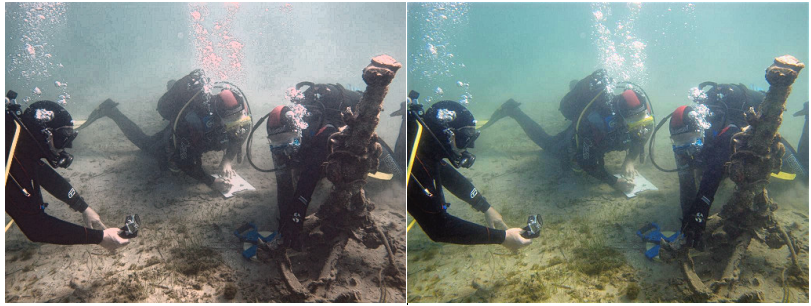

Fig. 3. The qualities of underwater images generated by different methods appears visually equal.

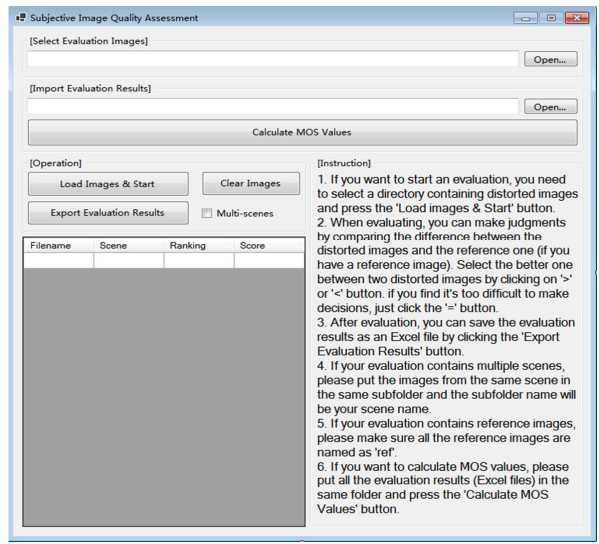

(a)

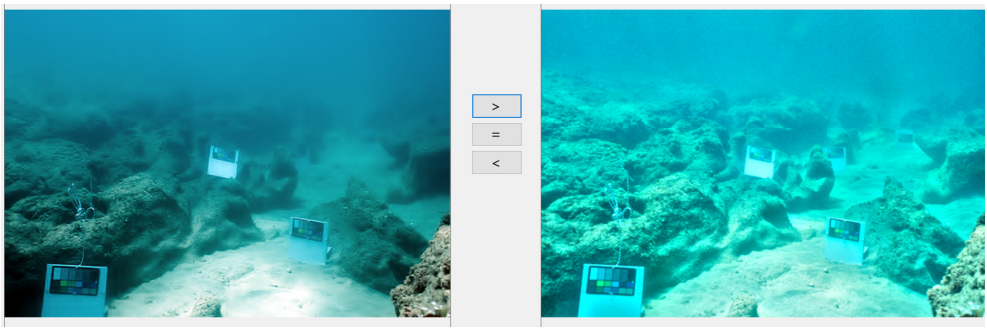

(b)

Fig. 4. Screenshot of evaluation software. (a) Home screen, (b) evaluation in progress.

Since the quality of enhanced or restored images by some different methods is visually equal, which is hard to decide which one is better, as shown in Fig. 3. In addition, the scale of our constructed UID2021 is regular. Therefore, in our subjective evaluation, we adopt the PCS scheme to conduct the quality comparison. We perform the subjective evaluation in an isolated room with natural illumination. In all sessions, a 21.5-inch LED Lenovo monitor at the solution of 1920×1080 pixels displays the two compared images. Observers are seated in front of the monitor at a distance of three times the screen height.

### 4.4 Methodology

A total of 52 undergraduate and graduate students participate in the subjective evaluation experiment. For improving the efficiency and convenience of rating and collecting, we design a subjective evaluation software, as shown in Fig. 4. Before the experiment, the participants are trained on using this software. Underwater images usually have practical applications, therefore, to make sure that subjects are not affected by any aesthetic factors in the process of subjective quality assessment, subjects are told to evaluate image quality from five aspects including color distortion, contrast distortion, texture distortion, visibility, recognizable foreground (see Table 5 for more details). In the experiment, all the 960 images are divided into six image subsets according to their degradation types. In each subset, observers can load the images with one scene or more than one scene at a time. Moreover, the software is programmed to ensure that the observers evaluate all images of the same scene before evaluating images of the other scenes. Besides, observers are instructed to rate each two compared images within no more than 10 seconds. In order to improve the shortcomings of the original PCS algorithm, it should be pointed out that when there are more than two pairs of images with equal quality, the system will automatically ask observers to rate them again

### 4.5 Data Processing

After using PCS in subjective evaluation, we obtain an integer sequence that presents the quality ranking of the images. To convert the ranking integers obtained from PCS into scores, we use the following formula to normalize these integers into [0, 9]:

$$Score_N(i) = 9 \times \frac{\max(r) - r(i)}{\max(r) - \min(r)} \tag{1}$$

where $r$ is the vector of ranking integers obtained from PCS, $i$ is the number of images contained in one scene $i = 1,2,\ldots,16$, and $Score_N(i)$ is the score of $i^{th}$ image rated by $N^{th}$ observer.

Evaluation results of image quality by different subjects usually contain abnormal results, which may be caused by wrong clicks, visual fatigue, and even depressed mood. These abnormal results will seriously contaminate the accuracy of our dataset. To prevent this, outlier detection and subject rejection are conducted with the same method in LIVE [17]. Approximately 3.77% of abnormal results and 1 subject have been removed from further consideration, and MOS values are calculated by averaging the left reliable individual scores.

### 4.6 Analysis of The Subjective Evaluation Results

As suggested in [59], the distribution of MOS and standard deviation of the subjective ratings are commonly adopted to verify the accuracy and suitability of the proposed dataset.

The distribution of MOS is indicative of the image quality range of a dataset. Generally, a uniform distribution of MOS is highly desired because it demonstrates that the range of subjective rating scales is completely utilized. In Fig. 5, we present the scatter plots of MOS distributions of our proposed dataset and UWIQA dataset, respectively. Fig. 6 shows the MOS distribution of six subsets of our UID2021. As shown in Fig. 5, we can observe that the MOS values of the UWIQA dataset are mostly located between 0.3 and 0.7, indicating that UWIQA covers a narrow range of image quality. On the contrary, the MOS distribution of our constructed UID2021 is very close to the uniform distribution, visually indicating that our UID2021 contains equal frequencies of images varying different levels of quality. Moreover, in Fig. 6, the MOS distributions of six subsets are also uniform, demonstrating that our proposed dataset fully utilizes the entire MOS range without emphasizing any part of that range.

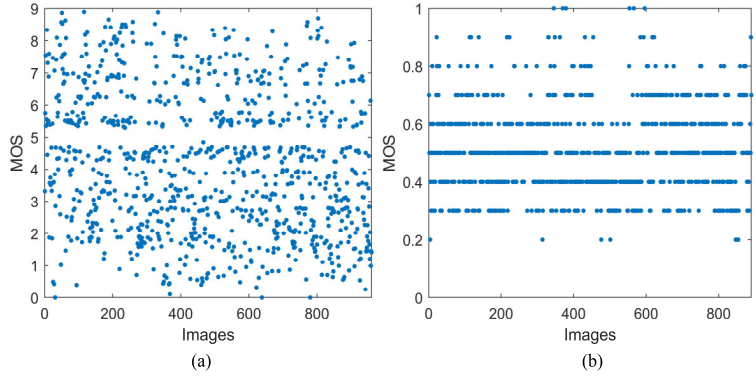

Fig. 5. Comparison of MOS distributions between (a) entire UID2021, and (b) UWIQA.

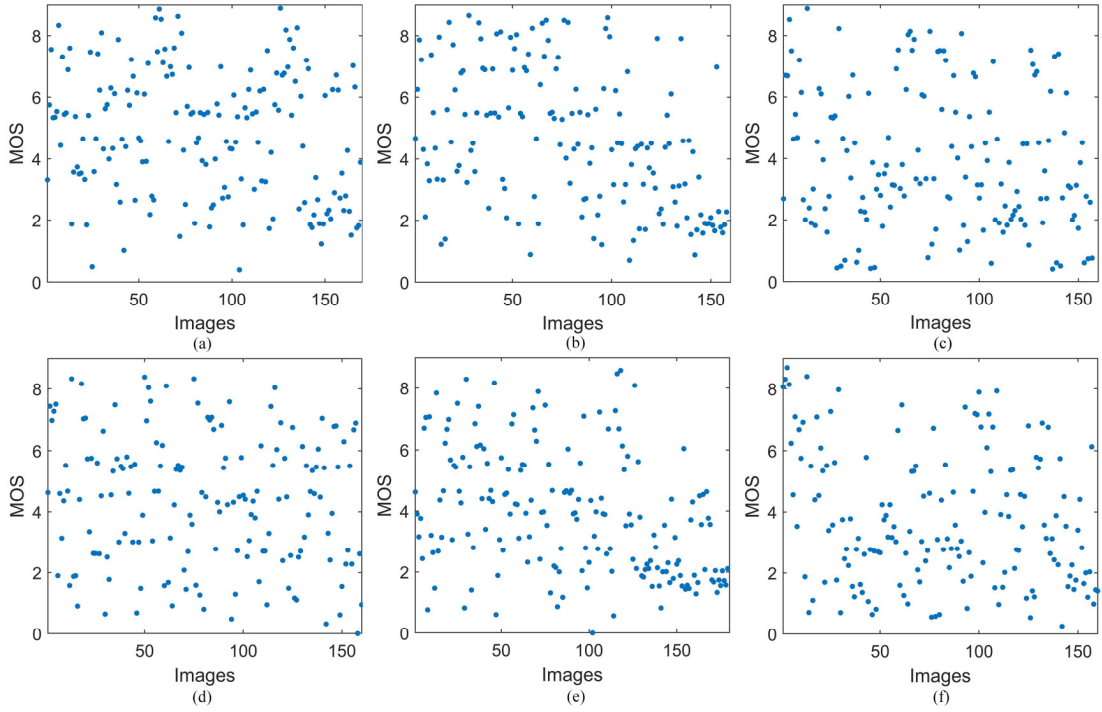

Fig. 6. The MOS distributions of six subsets with different scenes. (a) Blue scene, (b) blue-green scene, (c) green scene, (d) hazy scene, (e) low-light scene, and (f) turbid scene, respectively.

Table 6. The mean of standard deviations for "Bad", "Middle", and "Good" groups.

| Group | Bad | Middle | Good |
|---|---|---|---|
| Mean of Standard deviation | 1.1230 | 1.5009 | 1.1256 |

Table 7. Details of the NR IQA algorithms tested on our UID2021.

| IQA metrics | Description | Public implementation code |
|---|---|---|
| BIQI | A learning based model which trained on LIVE database. | http://live.ece.utexas.edu/research/quality/BIQI_release.zip |
| CPBD | A sharpness distortion-specific metric | http://ivulab.asu.edu/software |
| BLIIDNS-II | An efficient general-purpose blind image quality assessment (IQA) algorithm using a NSS model of discrete cosine transform coefficients | http://live.ece.utexas.edu/research/quality/BLIINDS2_release.zip |
| NIQE | A learning based approach. LIVE has been used for training. | http://live.ece.utexas.edu/research/quality/niqe.zip |
| FADE | A referenceless perceptual fog density prediction model | http://live.ece.utexas.edu/research/fog/index.html |
| UCIQE | Underwater IQA algorithm based on chroma, contrast, and saturation | https://github.com/paulwong16/UCIQE |
| UIQM | Underwater IQA algorithm based on human visual system | https://karen-panetta.squarespace.com/download |
| CCF | Underwater IQA algorithm based on contrast, colorfulness, and fog density | https://github.com/zhenglab/CCF |
| FDUM | A reference-free underwater image quality assessment metric in frequency domain | https://github.com/rmcong/Code-for-FDUM-method |

The standard deviation is another important benchmark. Here, in order to compare the MOS values with single individual scores, we further carry out a specific analysis of standard deviation by calculating the standard deviation of subjective scores on every single image in our UID2021, and obtained an average standard deviation of 1.2498. To further analyze the standard deviation, we divide all 960 images into three groups according to their corresponding MOS. The first group is called "Bad" with MOS from 0 to 3, followed by the second group called "Middle" with MOS from 3 to 6. Likewise, the last group is called "Good" with MOS larger than 6. Then, we calculate the mean of standard deviations separately for each group, and their results are present in Table 6. As we can see from Table 6, the standard deviation is typically higher in the "Middle" group than in the "Bad" and "Good" groups. To our opinion, this phenomenon means that subjects' opinions are much unified when evaluating images with bad or good quality, but they differ greatly when evaluating images with middle quality. All in all, the uniform distribution of MOS and a small standard deviation demonstrate that subjective evaluation in our proposed dataset is reliable.

## 5 EVALUATIONS

Since there is no available ground truth for real underwater images, UIQA databases are specially developed for testing the performance of different NR IQA algorithms. In this section, in order to fully understand the performance of the current NR metrics for IQA research, we perform a comprehensive evaluation of various widely used NR IQA metrics on our proposed dataset UID2021.

## 5.1 Criterion to Evaluate NR IQA Metrics

The four most widely used criteria for evaluating algorithms include Pearson linear correlation coefficient (PLCC), Spearman rank order correlation coefficient (SROCC), Kendall rank order correlation coefficient (KROCC), and root mean squared error (RMSE). Generally, the higher PLCC, SROCC, and KROCC as well as smaller RMSE indicate a better performance of the NR IQA algorithm. Specifically, before calculating PLCC and RMSE, the results of NR IQA algorithms are fitted to MOS through a five-parameter logistic function, which is defined by:

$$f(s) = \beta_1 \left( \frac{1}{2} - \frac{1}{1 + \exp(\beta_2 \cdot (s - \beta_3))} \right) + \beta_4 \cdot s + \beta_5 \quad (2)$$

where $s$ is the results of NR IQA algorithms, $f(s)$ is the fitted scores after nonlinear regression, and $\beta_i (i = 1, 2, \ldots, 5)$ are the regression function parameters which are estimated through the nonlinear fitting.

## 5.2 Comparison Results

Without reference images, researchers focus on exploring NR IQA metrics to evaluate underwater image quality. To the best of our knowledge, many NR IQA metrics have been used for testing the performance of underwater image enhancement and restoration algorithms [2, 47, 50, 52, 55-57, 60-65] including four underwater-specific NR-IQA algorithms in terms of UCIQE [4], UIQM [5], CCF [6], FDUM [7], and five in-air NR-IQA algorithms with regard to BIQI [30], CPBD [38], BLIINDS-II [33], NIQE [32] and FADE [40] (see Table 7 for details). However, there is no evidence to demonstrate whether they are suitable for underwater image quality assessment or not. Here, we further test these popular NR algorithms on our UID2021 database. Unlike in-air images databases, to provide valuable guidance, we first severally perform a more reasonable evaluation on the six subsets with different degradation scenes. The performance comparison of these nine NR IQA algorithms on the six common degradation scenes is successively presented in Table 8. The boldface values indicate the algorithms with the best performance in each scene.

From Table 8, we can see that compared with the underwater-specific NR IQA algorithms, the performances of tested in-air NR IQA algorithms are much poorer. To be specific, for the six scenes, UCIQE achieves the best predictive performance with respect to four criterion, closely followed by FDUM and UIQM. As for in-air NR IQA algorithms, FADE shows the best overall performance, followed by NIQE. However, we see that although these two algorithms perform well for turbid and low-light scenes, their performances are much worse for the blue-green and green scenes. For example, in low-light scene, the SROCC value of FADE reaches up to 0.5254, while in blue-green scene, the value is only about 0.16. Moreover, even the best algorithm, FADE, performs far from satisfactory, not mention to BIQI, CPBD, and BLIINDS-II. For the turbid scene, all four underwater-specific NR IQA algorithms achieve rather good performances with the highest SROCC being 0.7156. However, the SROCC of CCF drops considerably in the blue-green subset. Even the BIQI, an in-airs NR algorithm, performs better than CCF. To conclude, UCIQE and FDUM are very similar in performance, while FDUM performing better in the bluish scene and UCIQE performing better in the hazy scene. Moreover, UIQM is more suitable for the turbid scene, CCF performs worst among the four underwater NR algorithms. More unfortunately, all five in-air NR IQA algorithms can not get satisfactory results. It can be concluded that underwater images quality assessment is very challenging for in-air NR IQA algorithms and developing efficient NR IQA algorithms specific for underwater images is highly desired.

Additionally, we also conduct a performance comparison on the entire UID2021 dataset. Detailed results can also be found in Table 9. The conclusion is similar to those in Table 8, with FDUM and UCIQE demonstrating the best and second-best performances, followed by UIQM, CCF, and the other in-air IQA algorithms. To visually demonstrate the performances of these NR algorithms, we draw their scatter plots on our UID2021, as presented in Fig. 7. Here, we normalize the MOS values and IQA algorithms predicted scores into [0, 1] for better comparison and the blue lines are the ideal lines (MOS values equal to predicted scores). It also needs to be pointed out that for BIQI, FADE, BLIINDS-II, and FADE, the lower values indicate better image quality. In Fig. 7, it is clear to see that FDUM and UCIQE are closer to the ideal line. Additionally, in-air IQA algorithms in terms of BIQI, FADE, BLIINDS-II and NIQE tend to regard images with low MOS values as high-quality images. Considering CPBD, for images with MOS between 0 and 0.4, the results are generally around 0.7. Based on Fig. 7 and Table 9, we can also conclude that the performances of underwater NR algorithms are far superior to in-air NR algorithms.

Table 8. Performance comparison of popular NR IQA algorithms on six subsets.

| Degradation Scene | Criteria | In-air IQA algorithms | | | | | Underwater-specific IQA algorithms | | | |
|---|---|---|---|---|---|---|---|---|---|---|
| | | BIQI | CPBD | BLIINDS-II | NIQE | FADE | UCIQE | UIQM | CCF | FDUM |
| Blueish | SROCC | 0.1714 | 0.0032 | 0.2212 | 0.2828 | 0.3766 | 0.6182 | 0.5393 | 0.4570 | **0.6767** |
| | PLCC | 0.2730 | 0.2343 | 0.3277 | 0.3741 | 0.4109 | 0.6409 | 0.5908 | 0.5416 | **0.6862** |
| | KROCC | 0.1211 | 0.0042 | 0.1532 | 0.1953 | 0.2558 | 0.4575 | 0.3753 | 0.3281 | **0.5047** |
| | RMSE | 2.0238 | 2.0451 | 1.9875 | 1.9509 | 1.9179 | 1.6149 | 1.6972 | 1.7685 | **1.5302** |
| Blue-green | SROCC | 0.3980 | 0.1320 | 0.1611 | 0.3224 | 0.1547 | 0.5434 | 0.5042 | 0.2198 | **0.5513** |
| | PLCC | 0.4303 | 0.2224 | 0.2127 | 0.2399 | 0.2562 | 0.5767 | 0.5792 | 0.2699 | **0.5851** |
| | KROCC | 0.2686 | 0.0877 | 0.1126 | 0.2188 | 0.1015 | 0.3820 | 0.3531 | 0.1510 | **0.3889** |
| | RMSE | 1.9902 | 2.1495 | 2.1543 | 2.1403 | 2.1311 | 1.8012 | 1.7973 | 2.1229 | **1.7880** |
| Greenish | SROCC | 0.3345 | 0.0859 | 0.1028 | 0.2507 | 0.2890 | **0.6668** | 0.5864 | 0.4043 | 0.6588 |
| | PLCC | 0.3533 | 0.2598 | 0.1363 | 0.2485 | 0.2804 | **0.6905** | 0.6220 | 0.3097 | 0.6716 |
| | KROCC | 0.2297 | 0.0467 | 0.0711 | 0.1692 | 0.1898 | **0.4892** | 0.4164 | 0.2945 | 0.4783 |
| | RMSE | 2.1080 | 2.1748 | 2.2311 | 2.1815 | 2.1618 | **1.6290** | 1.7634 | 2.1414 | 1.6686 |
| Hazy | SROCC | 0.2821 | 0.027 | 0.0231 | 0.3442 | 0.364 | **0.6237** | 0.4291 | 0.4899 | 0.5261 |
| | PLCC | 0.3078 | 0.1504 | 0.1541 | 0.4036 | 0.4433 | **0.6688** | 0.5761 | 0.5668 | 0.5795 |
| | KROCC | 0.1933 | 0.0208 | 0.017 | 0.2352 | 0.2507 | **0.4558** | 0.3103 | 0.3496 | 0.379 |
| | RMSE | 2.0242 | 2.1034 | 2.1021 | 1.9466 | 1.9071 | **1.5817** | 1.739 | 1.7528 | 1.7339 |
| Low light | SROCC | 0.183 | 0.1655 | 0.1767 | 0.2618 | 0.5254 | **0.665** | 0.4928 | 0.434 | 0.62 |
| | PLCC | 0.2878 | 0.162 | 0.191 | 0.2947 | 0.5278 | **0.6727** | 0.5297 | 0.451 | 0.6261 |
| | KROCC | 0.1223 | 0.1127 | 0.1132 | 0.1775 | 0.3421 | **0.4755** | 0.3356 | 0.2978 | 0.439 |
| | RMSE | 1.9163 | 1.9468 | 1.9366 | 1.8853 | 1.7121 | **1.4597** | 1.6734 | 1.7609 | 1.5383 |
| Turbid | SROCC | 0.1982 | 0.0804 | 0.2826 | 0.5473 | 0.3897 | 0.5599 | **0.7156** | 0.6089 | 0.6725 |
| | PLCC | 0.3104 | 0.0881 | 0.4101 | 0.4093 | 0.3926 | 0.5844 | **0.7145** | 0.6383 | 0.6882 |
| | KROCC | 0.1366 | 0.0536 | 0.1962 | 0.3763 | 0.2693 | 0.4060 | **0.5357** | 0.4508 | 0.4987 |
| | RMSE | 2.0400 | 2.1377 | 1.9573 | 1.9580 | 1.9737 | 1.7414 | **1.5015** | 1.6520 | 1.5571 |

Table 9. The overall performance comparison of popular NR IQA algorithms on entire UID2021.

| Criteria | In-air IQA algorithms | | | | | Underwater-specific IQA algorithms | | | |
|---|---|---|---|---|---|---|---|---|---|
| | BIQI | CPBD | BLIINDS-II | NIQE | FADE | UCIQE | UIQM | CCF | FDUM |
| SROCC | 0.2605 | 0.0637 | 0.1077 | 0.3304 | 0.3835 | 0.6030 | 0.5404 | 0.4358 | **0.6269** |
| PLCC | 0.2725 | 0.1873 | 0.1185 | 0.3384 | 0.3905 | 0.6254 | 0.5816 | 0.4713 | **0.6374** |
| KROCC | 0.1789 | 0.0426 | 0.0721 | 0.2219 | 0.2588 | 0.4353 | 0.3807 | 0.3097 | **0.4515** |
| RMSE | 2.0924 | 2.1362 | 2.1594 | 2.0464 | 1.9993 | 1.6969 | 1.7691 | 1.9159 | **1.6756** |

To further confirm our judgments, we test the statistical significance of the performance of these NR IQA algorithms. The experimental results are tabulated in Table 10, where the values of "1", "0", and "-" indicate that the algorithm in the row is statistically superior, comparative, and inferior to the one in the column with 95% confidence. With the data in

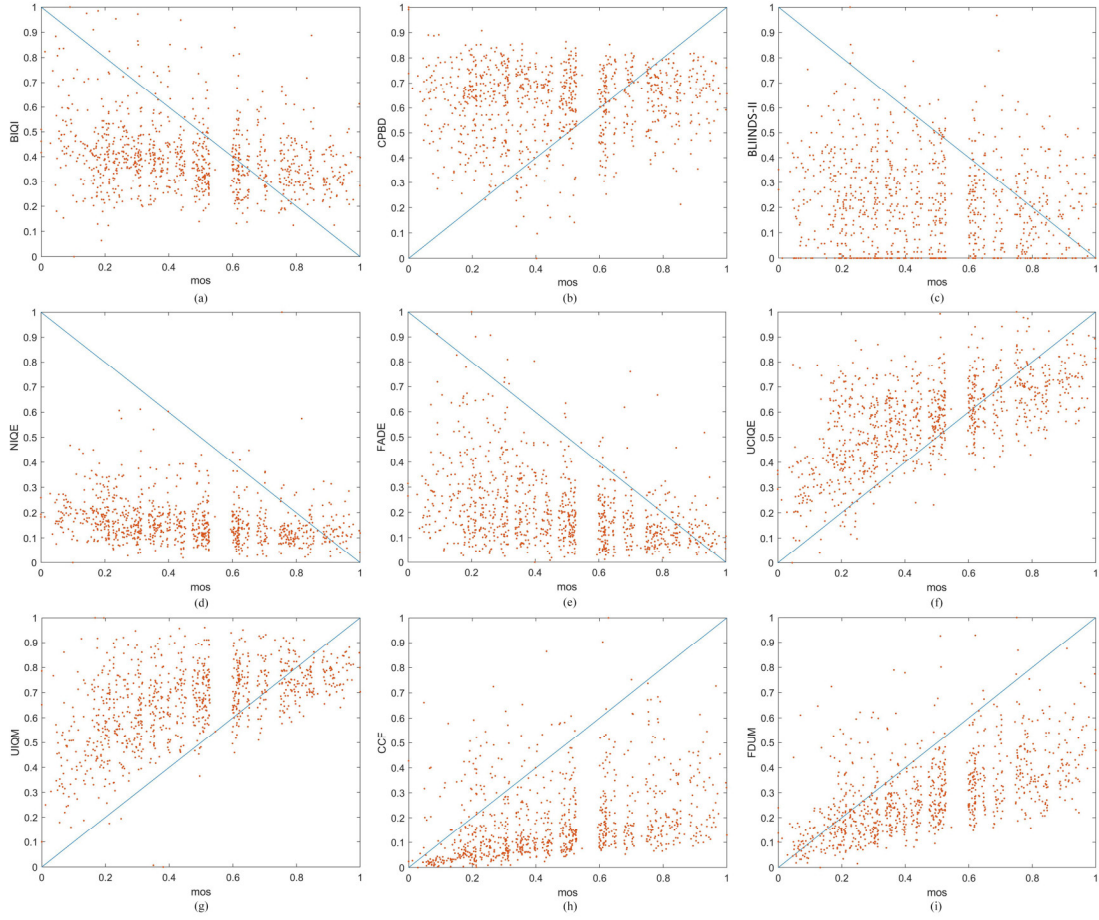

Fig. 7. Comparisons of the scatter plots of (a) BIQI, (b) CPBD, (c) BLIINDS-II, (d) NIQE, (e) FADE, (f) UCIQE, (g) UIQM, (h) CCF, (i) FDUM.

Table 10. Statistical significant analysis of NR IQA algorithms.

|  | BIQI | CPBD | BLIINDS-II | NIQE | FADE | UCIQE | UIQM | CCF | FDUM |
|---|---|---|---|---|---|---|---|---|---|
| BIQI | - | 1 | 1 | 0 | 0 | 0 | 0 | 0 | 0 |
| CPBD | 0 | - | 0 | 0 | 0 | 0 | 0 | 0 | 0 |
| BLIINDS-II | 0 | 1 | - | 0 | 0 | 0 | 0 | 0 | 0 |
| NIQE | 1 | 1 | 1 | - | 0 | 0 | 0 | 0 | 0 |
| FADE | 1 | 1 | 1 | 1 | - | 0 | 0 | 0 | 0 |
| UCIQE | 1 | 1 | 1 | 1 | 1 | - | 1 | 1 | - |
| UIQM | 1 | 1 | 1 | 1 | 1 | 0 | - | 1 | 0 |
| CCF | 1 | 1 | 1 | 1 | 1 | 0 | 0 | - | 0 |
| FDUM | 1 | 1 | 1 | 1 | 1 | - | 1 | 1 | - |

Table 10, we can observe that UCIQE and FDUM are statistically superior to all the other competitors on our dataset, demonstrating their superiority in quantifying the image quality.

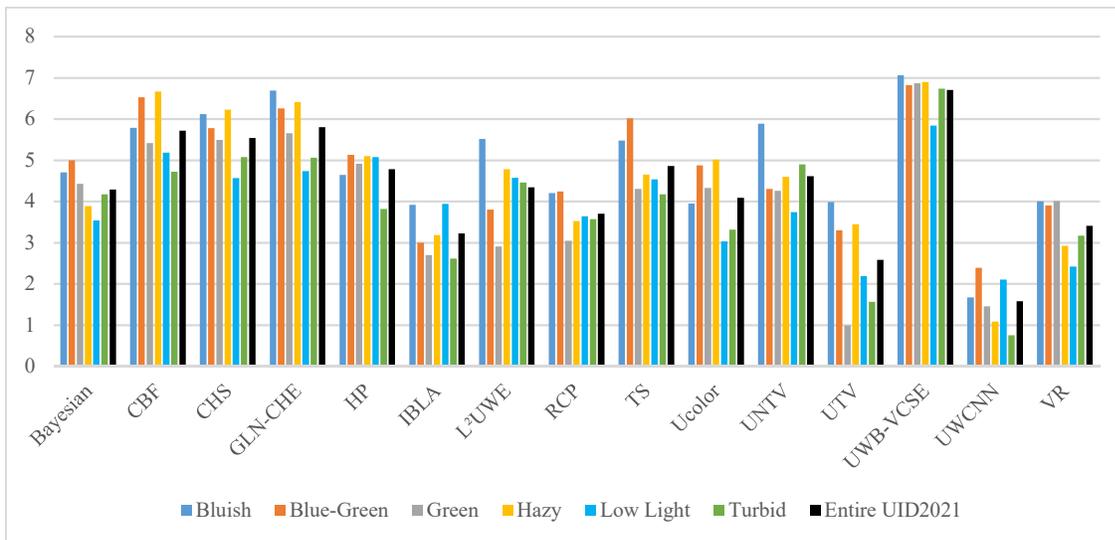

Fig. 8. Comparison of the average of MOS of different underwater enhancement and restoration methods.

Table 11. Comparison of the standard deviation of MOS of different underwater enhancement and restoration methods.

| Method | Standard deviation of MOS |
| --- | --- |
| Bayesian-retinex | 1.6678 |
| CBF | 1.9337 |
| CHS | 1.8847 |
| GLN-CHE | 2.0706 |
| HP | 1.9016 |
| IBLA | 1.5096 |
| $L^2$UWE | 1.7892 |
| RCP | 1.2669 |
| TS | 1.8830 |
| Ucolor | 1.4602 |
| UNTV | 1.8470 |
| UTV | 1.8188 |
| UWB-VCSE | 1.7062 |
| UWCNN | 0.9574 |
| VR | 1.6086 |

## 5.3 Evaluation of Underwater Image Enhancement and Restoration Methods

Since the corresponding 900 enhanced images are generated from 60 source images by employing fifteen underwater enhancement and restoration algorithms, it is also worthwhile to quantitatively analyze the performances of these algorithms. Following, we calculate the average MOS values of each method on six subsets and the full UID2021, respectively, and their results are given in Fig. 8. By comparing the average MOS values of different UIER algorithms in Fig. 8, we can conclude that UWB-VCSE, CBF, and GLN-CHE are the relatively best performers, which proves that the enhanced results of these algorithms are more in line with the human visual perception. Moreover, the limitation of lacking training data of UWCNN leads to producing some unsatisfactory results. Additionally, to further analyze the stability of these fifteen UIER methods, the standard deviations of MOS for each method are also calculated and their results are tabulated in Table 11. By jointly analyzing Fig. 8 and Table 11, it's interesting to note that the performances of several UIER algorithms fluctuated under different scenes, meaning that the current UIER algorithms are unable to generalize to all types of underwater images due to the complicated underwater environment and lighting conditions.

## 6 CONCLUSION

In this paper, we construct a new large-scale underwater image dataset, which is highly desired in the field of UIQA. Our proposed dataset contains 900 quality improved images that are derived from 60 real underwater images by utilizing fifteen SOTA underwater enhancement and restoration algorithms, generating 960 underwater images with different quality levels in total. The proposed dataset covers a diverse set of common underwater scenes, including blueish scene, greenish scene, blue-green scene, hazy scene, turbid scene, and low light scene. Both underwater-specific and in-air NR-IQA algorithms are tested on our dataset, and their experimental results suggest that evaluating the qualities of underwater images is very challenging for in-air IQA algorithms, and indicate the necessity of developing IQA algorithms specific for underwater images. Our constructed UID2021 dataset not only complements the existing underwater image datasets as a tool for testing the performance of objective NR IQA metrics, but also creates a platform for evaluating and optimizing the accuracy of newly upcoming UIQA metrics.